%% file: main.tex
\documentclass{article}

\PassOptionsToPackage{numbers, sort}{natbib}
\usepackage[final]{neurips_2024}
\usepackage[utf8]{inputenc} 
\usepackage[T1]{fontenc}   
\usepackage[colorlinks=true, allcolors=red]{hyperref}   
\usepackage{graphicx}
\usepackage{url}           
\usepackage{booktabs}      
\usepackage{amsfonts}     
\usepackage{amsmath}
\usepackage{nicefrac}       
\usepackage{microtype}      
\usepackage{xcolor}       
\usepackage{algorithm}
\usepackage{algpseudocode}
\usepackage{bm} 
\usepackage{cleveref}

\usepackage{lipsum}   
\usepackage{todonotes}

\title{Bayesian Predictive Coding}

\author{
\textbf{Alexander Tschantz}$^{1,2}$ \quad
\textbf{Magnus Koudahl}$^{1}$ \quad
\textbf{Hampus Linander}$^{1}$ \quad
\textbf{Lancelot Da Costa}$^{1}$ \\
\textbf{Conor Heins}$^{1,3}$ \quad
\textbf{Jeff Beck}$^{4}$ \quad
\textbf{Christopher Buckley}$^{1,2}$ \\
\\
$^1$VERSES AI Research Lab, Los Angeles, CA, USA \\
$^2$School of Engineering and Informatics, University of Sussex, Brighton, UK \\
$^3$Max Planck Institute of Animal Behavior, Department of Collective Behaviour, Konstanz, Germany \\
$^4$Department of Neurobiology, Duke University, Durham, NC, USA \\
\\
\texttt{\{alec.tschantz, magnus.koudahl, hampus.linander, lance.dacosta,} \\
\texttt{conor.heins, christopher.buckley\}@verses.ai, jeff.beck@duke.edu}
}
\begin{document}

\maketitle

\begin{abstract}
Predictive coding (PC) is an influential theory of information processing in the brain, providing a biologically plausible alternative to backpropagation. It is motivated in terms of Bayesian inference, as hidden states and parameters are optimised via gradient descent on variational free energy. However, implementations of PC rely on maximum \textit{a posteriori} (MAP) estimates of hidden states and maximum likelihood (ML) estimates of parameters, limiting their ability to quantify epistemic uncertainty. In this work, we investigate a Bayesian extension to PC that estimates a posterior distribution over network parameters. This approach, termed Bayesian Predictive coding (BPC), preserves the locality of PC and results in closed-form Hebbian weight updates. Compared to PC, our BPC algorithm converges in fewer epochs in the full-batch setting and remains competitive in the mini-batch setting. Additionally, we demonstrate that BPC offers uncertainty quantification comparable to existing methods in Bayesian deep learning, while also improving convergence properties. Together, these results suggest that BPC provides a biologically plausible method for Bayesian learning in the brain, as well as an attractive approach to uncertainty quantification in deep learning.
\end{abstract}

\section{Introduction}

Originating in the domain of neuroscience, the predictive coding (PC) framework \citep{rao1999predictive, mumford1992computational, friston2005theory, millidge2021predictive} proposes that the function of neural plasticity is to minimize local \textit{prediction errors}, which quantify the difference between estimated and observed signals. This framework has been adapted as a method for training deep neural networks using only local information, offering a biologically plausible alternative to backpropagation (BP) \citep{whittington2017approximation, millidge2022predictive, salvatori2022reverse, song2020can}. PC offers several benefits over BP, including improved performance in online and continual learning settings \citep{song2024inferring}, favourable optimization properties \citep{innocenti2024only}, the flexibility to be used in either a generative or discriminative manner \citep{millidge2022predictive2} and intrinsic auto-associative memory capabilities \citep{salvatori2021associative}. 

PC has traditionally been motivated through variational Bayesian inference \citep{buckley2017free, bogacz2017tutorial}, characterising the relationships between hidden states and parameters using probability distributions. However, implementations of PC generally do not operate on distributions but instead use maximum \textit{a posteriori} (MAP) estimates of hidden states and maximum likelihood (ML) estimates of parameters. This contrasts with Bayesian deep learning \citep{papamarkou2024position}, where the goal is to estimate posterior distributions over parameters so that epistemic and aleatoric uncertainty \citep{hullermeier2021aleatoric} can be quantified and subsequently used for model comparison, network pruning, or well-calibrated confidence estimation. Regardless, uncertainty quantification is essential for the robustness, reliability, and interpretability of learning systems, and how it is performed by the brain remains an open question. 

In this work, we propose an extension of PC that estimates approximate Bayesian posteriors over network parameters. This approach, termed Bayesian predictive coding (BPC), parameterises neural activity in a way that allows for the use of conjugate priors, enabling closed-form update rules for the weight distribution. The resulting updates are Hebbian functions of the pre- and post-synaptic activity, and hidden state updates retain their interpretation as precision-weighted prediction errors, thus preserving the locality and simplicity of the PC algorithm. Moreover, the ability to compute posterior updates in a closed form means that BPC can converge in substantially fewer iterations than gradient-based methods \citep{heins2024gradient}. See Appendix \ref{sec:related_work} for an overview of related work. 

In a set of experiments, we empirically verify that BPC achieves comparable performance to PC and traditional BP in full-batch training and remains competitive in mini-batch training. Notably, in the full-batch training context, BPC converges in a remarkably few epochs. Moreover, we demonstrate that the learned posterior distribution enables robust quantification of epistemic and aleatoric uncertainties in synthetic regression tasks. We compare BPC to a popular benchmark in Bayesian deep learning and demonstrate that our method provides both improved uncertainty quantification and improved accuracy and converges in fewer iterations. Taken together, our results suggest that BPC is a viable method for training uncertainty-aware neural networks using local information, suggesting a potential mechanism for uncertainty quantification in neural systems. 

\section{Methods}
\label{sec:methods}

Bayesian predictive coding is an algorithm for inverting hierarchical Gaussian generative models with \( L \) layers of variables \(\mathbf{Z} = \left\{\mathbf{z}_l: l= 0 \dots L\right\}\) and parameters \(\boldsymbol{\Theta} = \left\{\mathbf{W}_l, \boldsymbol{\Sigma}_l: l=1 \dots L\right\}\):
\begin{equation}   
    \begin{aligned}
           p(\mathbf{Z}, \boldsymbol{\Theta}) &= p(\mathbf{z}_0) p(\mathbf{W}_0, \boldsymbol{\Sigma}_0) \prod_{l=1}^{L} p(\mathbf{z}_l \mid \mathbf{z}_{l-1}, \mathbf{W}_l, \boldsymbol{\Sigma}_l) p(\mathbf{W}_l, \boldsymbol{\Sigma}_l), \\
           p(\mathbf{z}_l \mid \mathbf{z}_{l-1}, \mathbf{W}_l, \boldsymbol{\Sigma}_l) &= \mathcal{N}(\mathbf{z}_l \mid  \mathbf{W}_l f(\mathbf{z}_{l-1}), \boldsymbol{\Sigma}_l), \\
           p(\mathbf{W}_l, \boldsymbol{\Sigma}_l) &= \mathcal{MN}(\mathbf{W}_l \mid \mathbf{M}_l^{(0)}, \boldsymbol{\Sigma}^{-1}_l, \mathbf{V}_l^{(0)}) \ \mathcal{W}(\boldsymbol{\Sigma}_l^{-1} \mid \boldsymbol{\Psi}_l^{(0)}, \nu_l^{(0)}),
   \end{aligned}
   \label{eq:joint_distribution}
\end{equation}
where $f(\cdot)$ is a point-wise non-linear activation function, and \( p(\mathbf{z}_0) = \mathcal{N}(\mathbf{z}_0 \mid \boldsymbol{\mu}_0, \boldsymbol{\Sigma}_0) \). Notably, we have placed the parameters $\mathbf{W}_l$ outside of the non-linear activation function $f(\cdot)$, which is essential for enabling the closed-form updates used in Equation \ref{eq:update_natural_params}. In practice, we also use a bias term that we ignore here for simplicity. Given that the log-likelihood is quadratic in the network parameters, a Matrix Normal Wishart prior for parameters $\mathbf{W}_l$ and $\boldsymbol{\Sigma}_l$ is conditionally conjugate given network activity $\mathbf{Z}$. See Appendix \ref{sec:distributions} for relevant definitions.  

In the standard PC framework, latent variables \(\mathbf{Z}\) are represented via maximum a posteriori (MAP) estimates, and parameters $\boldsymbol{\Theta}$ are represented via maximum likelihood (ML) estimates. In contrast, BPC augments this approach by representing an approximate posterior distribution over parameters, 
\begin{equation}
\begin{aligned}
        q_{\boldsymbol{\lambda}}(\boldsymbol{\Theta}) &= \prod_{l=1}^{L} q(\mathbf{W}_l, \boldsymbol{\Sigma}_l), \\
        q_{\boldsymbol{\lambda}}(\mathbf{W}_l, \boldsymbol{\Sigma}_l) &= \mathcal{MN}(\mathbf{W}_l \mid \mathbf{M}_l, \boldsymbol{\Sigma}^{-1}_l, \mathbf{V}_l) \ \mathcal{W}(\boldsymbol{\Sigma}^{-1}_l \mid \boldsymbol{\Psi}_l, \nu_l).
\end{aligned}
\label{eq:approxpost_params}
\end{equation}
\noindent where $\boldsymbol{\lambda} = \left\{\mathbf{M}_l, \mathbf{V}_l, \boldsymbol{\Psi}_l, \nu_l, : l=1 \dots L\right\}$ are the variational parameters. Given Equation \eqref{eq:joint_distribution} and \eqref{eq:approxpost_params}, we define the objective function $\mathcal{E}(\mathbf{Z}, \boldsymbol{\lambda})$ as:
\begin{equation}
        \mathcal{E}(\mathbf{Z}, \boldsymbol{\lambda}) = \Big< \log q_{\boldsymbol{\lambda}}(\boldsymbol{\Theta}) - \log p(\mathbf{Z}, \boldsymbol{\Theta}) \Big>_{q_{\boldsymbol{\lambda}}(\boldsymbol{\Theta})}  
        \label{eq:energy_objective}
\end{equation}
which is equivalent to the variational free energy under the assumption that $q_\lambda(\mathbf{Z})$ is a Dirac delta distribution \cite{buckley2017free}. See Appendix \ref{sec:pc} comparing Equation \eqref{eq:energy_objective} to the energy function used in PC. To estimate \(\mathbf{Z}\) and \(\boldsymbol{\Theta}\), we use the expectation-maximisation (EM) algorithm \cite{dempster1977maximum} on \(\mathcal{E}(\mathbf{Z}, \boldsymbol{\lambda})\):
\begin{equation}
    \mathbf{Z}^* = \arg \min_{\mathbf{Z}} \mathcal{E}(\mathbf{Z}, \boldsymbol{\lambda}), \quad \boldsymbol{\lambda}^* = \arg \min_{\boldsymbol{\lambda}} \mathcal{E}(\mathbf{Z}^*, \boldsymbol{\lambda}).
    \label{eq:em_algorithm}
\end{equation}
Following PC, we solve the first optimisation problem using gradient descent on the energy function \(\mathcal{E}(\mathbf{Z}, \boldsymbol{\lambda})\), which can be rewritten as:
\begin{equation}
    \mathcal{E}(\mathbf{Z}, \boldsymbol{\lambda}) = \frac{1}{2} \sum_{l=1}^{L} \underbrace{\left<\left(\mathbf{z}_l - \mathbf{W}_l f(\mathbf{z}_{l-1}) \right)^\top \boldsymbol{\Sigma}_l^{-1} \left(\mathbf{z}_l - \mathbf{W}_l f(\mathbf{z}_{l-1}) \right)\right>_{ q_{\boldsymbol{\lambda}}(\mathbf{W}_l, \boldsymbol{\Sigma}_l)}}_{\mathcal{E}_l} + C,
    \label{eq:energy_rewrite}
\end{equation}
where $C$ represents terms independent of $\mathbf{Z}$ and $\mathcal{E}_l$ is the \textit{precision-weighted prediction error} for layer $l$. Gradient descent is performed by iteratively updating the latent variables \(\mathbf{Z}\) using the following update rule:
\begin{equation}
    \mathbf{z}_l \leftarrow \mathbf{z}_l - \alpha \frac{\partial \mathcal{E}}{\partial \mathbf{z}_l}, \quad  \frac{\partial \mathcal{E}}{\partial \mathbf{z}_l} = \frac{1}{2} \left( \frac{\partial \mathcal{E}_l}{\partial \mathbf{z}_l} + \frac{\partial \mathcal{E}_{l+1}}{\partial \mathbf{z}_l} \right),
    \label{eq:gradient_descent}
\end{equation}
where \(\alpha\) is the learning rate. This process is repeated until convergence or a maximum number of iterations \( T \), and the converged latent variables are denoted as \(\mathbf{Z}^{\star}\). See Appendix \ref{sec:derivatives} for the closed-form expressions of the derivatives in Equation \eqref{eq:gradient_descent} as well as a method for optimally selecting $\alpha$.

Given the converged latent variables \(\mathbf{Z}^{\star}\), we can find a closed-form expression for $\arg \min_{\boldsymbol{\lambda}} \mathcal{E}(\mathbf{Z}^*, \boldsymbol{\lambda})$. Specifically, we update the natural parameters \(\boldsymbol{\eta}_l\) of the variational posterior $ q_{\boldsymbol{\lambda}}(\mathbf{W}_l, \boldsymbol{\Sigma}_l)$:
\begin{equation}
    \boldsymbol{\eta}_l^{\star} = \boldsymbol{\eta}^{(0)}_l + \sum_n \left( f(\mathbf{z}^{*n}_{l-1}) f(\mathbf{z}^{*n}_{l-1})^\top, f(\mathbf{z}^{*n}_{l-1}){\mathbf{z}^{*n}_{l}}^\top, \mathbf{z}^{n*}_{l}{\mathbf{z}^{n*}_{l}}^\top, \mathbf{1}\right),
    \label{eq:update_natural_params}
\end{equation}
where and \(\boldsymbol{\eta}^{(0)}_l\) are the natural parameters for the prior distribution defined in Equation \eqref{eq:joint_distribution}, and $\mathbf{1}$ denotes a vector of ones whose dimension matches the number of data points $n$. Equation \ref{eq:update_natural_params} has a natural interpretation as a Hebbian function of pre- and post-synaptic activity. See Appendix  \ref{sec:closed_form} for a proof that Equation \ref{eq:update_natural_params} minimises $\mathcal{E}(\mathbf{Z}^*, \boldsymbol{\lambda})$. The natural parameters \(\boldsymbol{\eta}_l\) relate to the parameters of Equation \eqref{eq:approxpost_params} via:
\begin{equation}
    \boldsymbol{\eta}_l = \left( \mathbf{V}_l^{-1}, \, \mathbf{M}_l \mathbf{V}_l^{-1}, \, \mathbf{\Psi}_l^{-1} + \mathbf{M}_l \mathbf{V}_l^{-1} \mathbf{M}_l^\top, \, \nu_l - d_{y_l} + d_{x_l} - 1 \right).
    \label{eq:natural_params_relation}
\end{equation}
Equation \ref{eq:update_natural_params} provides an exact solution to the optimisation problem when the latent variables $\mathbf{Z}^\star$ are estimated for the entire dataset. In practice, mini-batching is often used, in which case we introduce a learning rate $\kappa$ and update the natural parameters as $\boldsymbol{\eta}_l^{\star} = (1 - \kappa) \boldsymbol{\eta}_l + \kappa  \boldsymbol{\eta}_l^{\star}$. This minibatch update rule corresponds to performing stochastic natural gradient descent on the variational parameters, leveraging the geometry induced by the variational posterior distribution \citep{hoffman2013stochastic}.

\begin{algorithm}
    \caption{Bayesian Predictive Coding}
    \begin{algorithmic}[1]
        \State  Randomly initialize $\boldsymbol{\eta}$ and set prior parameters $\boldsymbol{\eta}^{(0)} = (\mathbf{M}^{(0)} \mathbf{V}^{(0)} \boldsymbol{\Psi}^{(0)}, \nu^{(0)})$
    \For{each $(\mathbf{x}, \mathbf{y})$ batch}
        \Repeat 
            \State Initialize $\mathbf{Z}$
            \For{$l = 1$ to $L$}
                \State $\mathbf{z}_l \leftarrow \mathbf{z}_l - \alpha \frac{\partial \mathcal{E}}{\partial \mathbf{z}_l}$
            \EndFor
        \Until{convergence of $\mathbf{z}$ or maximum $T$ iterations}
        \For{$l = 1$ to $L$}
            \State $\boldsymbol{\eta}_l \leftarrow \boldsymbol{\eta}^{(0)}_l + \sum_n \left( f(\mathbf{z}^{*n}_{l-1}) f(\mathbf{z}^{*n}_{l-1})^\top, f(\mathbf{z}^{*n}_{l-1}){\mathbf{z}^{*n}_{l}}^\top, \mathbf{z}^{n*}_{l}{\mathbf{z}^{n*}_{l}}^\top, \mathbf{1}\right)$
        \EndFor
    \EndFor
    \end{algorithmic}
\label{alg:1}
\end{algorithm}

\paragraph{Training and testing}

Given a dataset \(\mathcal{D} = \left\{(\mathbf{x}^{(i)}, \mathbf{y}^{(i)})\right\}_{i=1}^N\), Bayesian Predictive Coding (BPC) trains models in a discriminative manner by fixing the input nodes to \(\mathbf{z}_0 = \mathbf{x}^{(i)}\) and the output nodes to \(\mathbf{z}_L = \mathbf{y}^{(i)}\). Alternatively, the model can also be trained in an unsupervised setting by only fixing the top layer \(\mathbf{z}_L\). For each mini-batch of data, we iteratively apply Equation \eqref{eq:em_algorithm}, as detailed in Algorithm \ref{alg:1}.

During testing, we handle the uncertainty captured by the parameter posterior \(q_{\boldsymbol{\lambda}}(\boldsymbol{\Theta})\) in three distinct ways:

\textbf{Deterministic forward pass}: We use the expected parameter values at each layer, effectively ignoring uncertainty:
\begin{equation}
    \left< \mathbf{W}_l, \boldsymbol{\Sigma}_l \right>_{q_{\boldsymbol{\lambda}}(\mathbf{W}_l, \boldsymbol{\Sigma}_l)}.
\end{equation}

\textbf{Monte Carlo sampling}: We sample parameters from their posterior distributions:
\begin{equation}
\mathbf{W}_l \sim \mathcal{MN}(\mathbf{W}_l \mid \mathbf{M}_l, \boldsymbol{\Sigma}^{-1}_l, \mathbf{V}_l), \quad \boldsymbol{\Sigma}_l^{-1} \sim \mathcal{W}(\boldsymbol{\Sigma}_l^{-1} \mid \boldsymbol{\Psi}_l, \nu_l),
\end{equation}
and average the network outputs across multiple samples to estimate predictive uncertainty.

\textbf{Analytical uncertainty propagation}: We analytically propagate the uncertainty through each layer of the network by computing:
\begin{equation}
    q(\mathbf{z}_l) \propto \Big< \mathcal{N}(\mathbf{z}_l \mid \mathbf{W}_l \mathbf{z}_{l-1}, \boldsymbol{\Sigma}_l) \Big>_{q(\mathbf{z}_{l-1}) q_{\boldsymbol{\lambda}}(\mathbf{W}_l, \boldsymbol{\Sigma}_l)},
\end{equation}
where \(q(\mathbf{z}_{l-1})\) is approximated as Gaussian (or Dirac delta for the input layer). The uncertainty is propagated through the nonlinear activation function \(f(\cdot)\) using the deterministic approximation method from \citep{wu2018deterministic}, suitable for the \texttt{ReLU} activations employed in our experiments.

\section{Experiments}

\subsection{Accuracy}
\label{sec:acc}

We implement the BPC algorithm to train \texttt{ReLU} networks and evaluate their accuracy to networks trained via PC and BP. Specifically, we examine performance on two small datasets (the \texttt{energy} dataset from the UCI repository \citep{Dua:2019} and the \texttt{two moons} dataset \cite{scikit-learn}) using full-batch training, and one larger dataset (\texttt{MNIST} \citep{deng2012mnist}) using mini-batch training. See \Cref{sec:experiment_details} for details on hyperparameters and dataset specifics.

\begin{figure}[t] 
    \centering
    \includegraphics[width=1.0\linewidth]{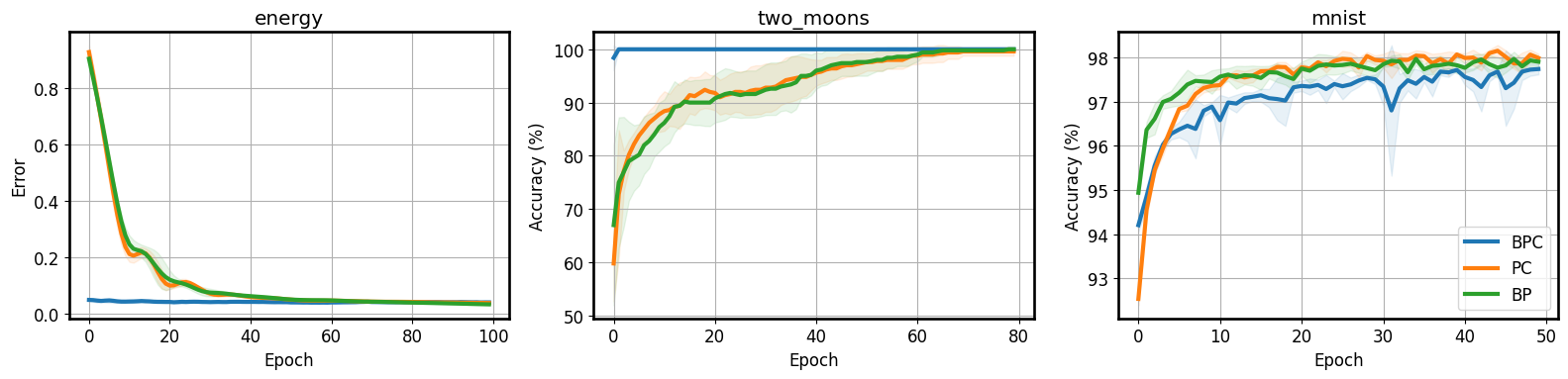}
    \caption{ Accuracy on two classification tasks (\texttt{two moons} and \texttt{MNIST}) and error on one regression task (\texttt{energy}). These results show that BPC converges to the same accuracy in fewer epochs in the full-batch setting (\texttt{energy} and \texttt{two moons}), and is competitive in the mini-batch setting (\texttt{MNIST}). Shaded regions denote 1 standard deviation across 5 seeds. See Appendix \ref{sec:experiment_details} for experiment details.\label{fig:1}}
\end{figure}

The results of these experiments are shown in \Cref{fig:1}. In the full-batch setting, BPC converges in the first few epochs, since it utilises closed-form updates for the parameter posterior, whereas PC and BP both take several epochs to converge. In the mini-batch setting, BPC is competitive with both PC and BP, converging to an accuracy that is, on average, within $0.3\%$ of these methods. We note that BP and PC are both optimised using the \texttt{Adam} optimiser \citep{kingma2014adam}, with PC additionally requiring weight decay \cite{kinghorn2022preventing}. When trained with vanilla stochastic gradient descent (SGD), BP and PC converge significantly slower than BPC, and the accuracy of PC is often significantly lower. These experiments confirm that the posterior updates in Equation \ref{eq:update_natural_params} provide a viable method for training Bayesian deep neural networks using local update rules.

\subsection{Uncertainty quantification}

To evaluate the learned posterior $q_{\boldsymbol{\lambda}}(\boldsymbol{\Theta})$, we train a compact BPC model on two synthetic regression tasks and empirically verify that the model can quantify aleatoric and epistemic uncertainty. To quantify aleatoric uncertainty, we propagate uncertainty through the network to estimate the first and second-order moments of the output. This approach naturally accommodates homoscedastic variance; for heteroscedastic variance, we additionally parameterise the output layer with a variance node following the method described in \citep{wu2018deterministic}. We quantify epistemic uncertainty by drawing multiple samples from the parameter posterior $q_{\boldsymbol{\lambda}}(\boldsymbol{\Theta})$ and visualizing the predicted functions. These results, illustrated in Figure \ref{fig:2}, confirm that our model accurately captures both forms of uncertainty. See Appendix \ref{sec:experiment_details} for details on the regression tasks and network hyperparameters used.

\begin{figure}[t] 
    \centering
    \includegraphics[width=1.0\linewidth]{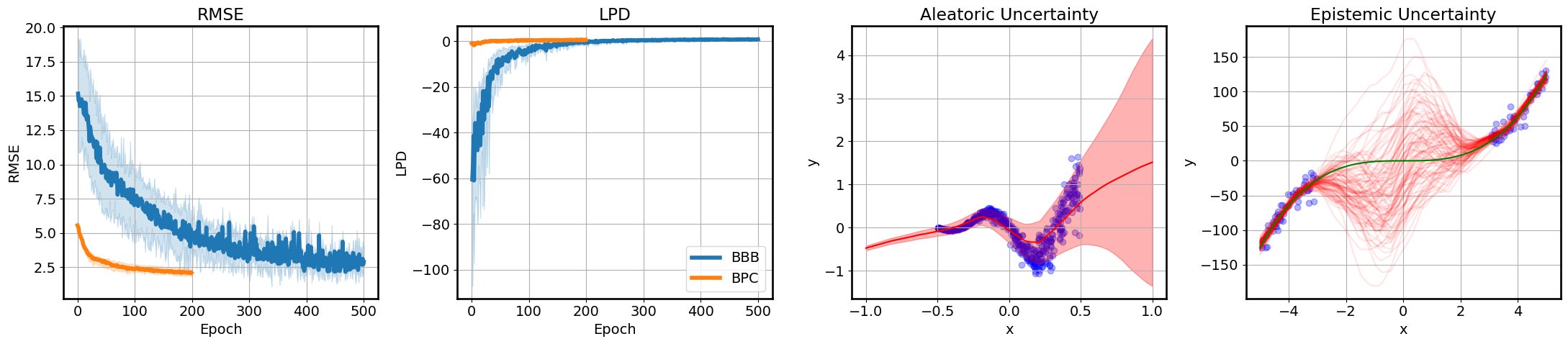} 
    \caption{Comparison of root mean square error (RMSE) and log predictive density (LPD) metrics on the UCI regression dataset \texttt{yacht} using Bayesian Predictive Coding (BPC) and Bayes by Backprop (BBB). The final two plots show the aleatoric and epistemic uncertainties quantified by the model on synthetic regression tasks. See Appendix \ref{sec:experiment_details} for experiment details.}
    \label{fig:2}
\end{figure}

Finally, we compare BPC to a popular Bayesian deep learning benchmark, \textit{Bayes by Backprop} (BBB) \citep{blundell2015weight}. BBB estimates variational free energy by drawing samples from the posterior distribution and subsequently uses backpropagation to update the variational parameters. We compare BPC and BBB using log predictive density (LPD) and root mean squared error (RMSE) across several UCI regression tasks \citep{Dua:2019}. For both methods, LPD is calculated by drawing multiple samples from the posterior distribution over weights and computing the average log-likelihood of the data points. Table \ref{tab:uci} shows that BPC outperforms BBB in terms of both LPD and RMSE on most tasks. Figure \ref{fig:2} plots LPD and RMSE against training epochs for the \texttt{yacht} dataset, demonstrating faster convergence of BPC due to its closed-form updates. We observe similar improvements in convergence across other datasets.

\begin{table}[ht]
\centering
\setlength{\tabcolsep}{10pt} 
\renewcommand{\arraystretch}{1.0} 
\begin{tabular}{l|rr|rr}
  \hline
  \texttt{Dataset} & \multicolumn{2}{c|}{\texttt{LPD}} & \multicolumn{2}{c}{\texttt{RMSE}} \\
  & \texttt{BBB} & \texttt{BPC} & \texttt{BBB} & \texttt{BPC} \\
  \hline
  \texttt{yacht}     & $\mathbf{1.02}$  & $0.75$           & $2.16$           & $\mathbf{2.08}$  \\ 
  \texttt{concrete}  & $-0.77$          & $\mathbf{-0.59}$ & $6.03$           & $\mathbf{5.60}$  \\ 
  \texttt{wine}      & $-8.93$          & $\mathbf{-2.49}$ & $0.64$           & $\mathbf{0.60}$  \\ 
  \texttt{housing}   & $-0.56$          & $\mathbf{-0.49}$ & $2.81$           & $\mathbf{2.62}$  \\ 
  \texttt{power}     & $-0.39$          & $\mathbf{-0.14}$ & $\mathbf{4.11}$  & $4.13$           \\ 
  \texttt{energy}    & $\mathbf{0.33}$           & $0.28$  & $1.86$           & $\mathbf{1.51}$  \\ 
  \hline
\end{tabular}
\vspace{5pt}
\caption{Comparison between Bayesian Predictive Coding (BPC) and Bayes by backprop (BBB) 
 in terms of the average log-likelihood (LPD) and root mean square error (RMSE) for UCI regression datasets. See Appendix \ref{sec:experiment_details} for experiment details.}
\label{tab:uci}
\end{table}

\section{Discussion}

In this work, we have introduced Bayesian predictive coding (BPC), an algorithm that extends predictive coding (PC) by incorporating Bayesian posterior distributions over model parameters. We demonstrated that the resulting update rules naturally translate into Hebbian functions of pre- and post-synaptic activity, thus preserving the local computations and biological plausibility central to PC. Additionally, these update rules provide closed-form expressions for posterior parameters, enabling enhanced convergence properties in full-batch training. Collectively, our findings suggest that BPC offers a viable approach for implementing Bayesian neural networks in biological systems.

The current work has two main limitations. First, it inherits the computational cost associated with performing gradient descent on latent variables $\mathbf{Z}$ before each weight update. This limitation is also shared by PC, resulting in similar computational times per epoch for both algorithms. Second, the use of the Matrix Normal Wishart distribution for parameter posteriors introduces additional computational complexity. For larger neural networks, structured low-rank approximations will become necessary, and the choice of posterior approximations is an important direction for future research. We note that all experiments presented here were executed on consumer-grade CPUs.

Several avenues for further exploration remain. For instance, it would be possible to take a model pre-trained using backpropagation and subsequently apply BPC to quantify model uncertainty on new data batches by estimating latent variables $\mathbf{Z}$ and performing closed-form posterior updates for $\boldsymbol{\lambda}$. Additionally, further research could investigate the optimization properties of BPC more thoroughly. For example, the current estimate of $\boldsymbol{\Sigma}$ acts as an adaptive learning rate during inference of latent variables $\mathbf{Z}$, dynamically emphasizing more informative (low-variance) dimensions. Furthermore, employing conjugate priors $p(\boldsymbol{\Theta})$ may promote beneficial optimization behavior; for instance, favoring identity-like priors for $\boldsymbol{\Sigma}$ may encourage independence among latent dimensions, potentially leading to more disentangled representations conducive to cross-task generalization.

\newpage
\bibliographystyle{unsrtnat}
\bibliography{main}

\newpage
\input{appendix.tex}

\end{document}

%% file: appendix.tex
\appendix

\section{Distributions}
\label{sec:distributions}

\paragraph{Matrix Normal Distribution}

The Matrix Normal distribution is a generalization of the multivariate normal distribution to matrix-valued random variables. A random matrix \(\mathbf{W} \in \mathbb{R}^{d_y \times d_x}\) is said to follow a Matrix Normal distribution with mean \(\mathbf{M} \in \mathbb{R}^{d_y \times d_x}\), row precision \(\boldsymbol{\Sigma}^{-1} \in \mathbb{R}^{d_y \times d_y}\), and column covariance \(\mathbf{V} \in \mathbb{R}^{d_x \times d_x}\), denoted as \(\mathcal{MN}(\mathbf{W} \mid \mathbf{M}, \boldsymbol{\Sigma}^{-1}, \mathbf{V})\), if its probability density function is given by:

\begin{equation}
    p(\mathbf{W} \mid \mathbf{M}, \boldsymbol{\Sigma}^{-1}, \mathbf{V}) = \frac{\exp\left(-\frac{1}{2} \mathrm{Tr}\left(\boldsymbol{\Sigma}^{-1}(\mathbf{W} - \mathbf{M}) \mathbf{V}^{-1}(\mathbf{W} - \mathbf{M})^\top\right)\right)}{(2\pi)^{\frac{d_y d_x}{2}} |\boldsymbol{\Sigma}|^{\frac{d_x}{2}} |\mathbf{V}|^{\frac{d_y}{2}}}.
\end{equation}

\paragraph{Wishart Distribution}

The Wishart distribution is a distribution over symmetric positive definite matrices and is the conjugate prior for the covariance matrix in multivariate Gaussian distributions. In our formulation, let \(\boldsymbol{\Sigma}^{-1}\) be a random matrix that follows a Wishart distribution with scale matrix \(\boldsymbol{\Psi}\) and degrees of freedom \(\nu\), denoted as \(\mathcal{W}(\boldsymbol{\Sigma}^{-1} \mid \boldsymbol{\Psi}, \nu)\). The density function is given by:

\begin{equation}
    p(\boldsymbol{\Sigma}^{-1} \mid \boldsymbol{\Psi}, \nu) = \frac{|\boldsymbol{\Sigma}^{-1}|^{\frac{\nu - d_y - 1}{2}} \exp\left(-\frac{1}{2} \mathrm{Tr}(\boldsymbol{\Sigma}^{-1} \boldsymbol{\Psi}^{-1})\right)}{2^{\frac{\nu d_y}{2}} |\boldsymbol{\Psi}|^{\frac{\nu}{2}} \Gamma_{d_y}\left(\frac{\nu}{2}\right)},
\end{equation}

where \(\Gamma_{d_y}(\cdot)\) denotes the multivariate gamma function.

\paragraph{Matrix Normal Wishart Distribution}

The Matrix Normal Wishart distribution is the joint distribution of \(\mathbf{W}\) and \(\boldsymbol{\Sigma}\). If \(\mathbf{W} \mid \boldsymbol{\Sigma}^{-1} \sim \mathcal{MN}(\mathbf{W} \mid \mathbf{M}, \boldsymbol{\Sigma}^{-1}, \mathbf{V})\) and \(\boldsymbol{\Sigma}^{-1} \sim \mathcal{W}(\boldsymbol{\Sigma}^{-1} \mid \boldsymbol{\Psi}, \nu)\), then the joint density is given by:

\begin{equation}
\begin{aligned}
    \log p(\mathbf{W}, \boldsymbol{\Sigma} \mid \mathbf{M}, \mathbf{V}, \boldsymbol{\Psi}, \nu) &= -\frac{1}{2} \mathrm{Tr}\left(\boldsymbol{\Sigma}^{-1}(\mathbf{W} - \mathbf{M})\mathbf{V}^{-1}(\mathbf{W} - \mathbf{M})^\top\right) + \frac{d_x}{2} \log |\boldsymbol{\Sigma}^{-1}| \\
    &\quad -\frac{1}{2} \mathrm{Tr}(\boldsymbol{\Sigma}^{-1} \boldsymbol{\Psi}^{-1}) + \frac{\nu - d_y - 1}{2} \log |\boldsymbol{\Sigma}^{-1}| - \frac{d_y d_x}{2} \log(2\pi) \\
    &\quad + \frac{d_y}{2} \log |\mathbf{V}^{-1}| - \frac{\nu d_y}{2} \log(2) - \log \Gamma_{d_y}\left(\frac{\nu}{2}\right).
\end{aligned}
\end{equation}

Here, \(\mathbf{M}\) is the mean matrix, \(\mathbf{V}\) is the column covariance matrix, \(\boldsymbol{\Psi}\) is the scale matrix, and \(\nu\) is the degrees of freedom.

\section{Derivatives}
\label{sec:derivatives}

The derivatives of the energy function from Equation \eqref{eq:gradient_descent} are given by:

\begin{align}
\nabla_{\mathbf{z}_{l}} \mathcal{E}_l &= \left<\boldsymbol{\Sigma}_l^{-1} (\mathbf{z}_l - \mathbf{W}_l f(\mathbf{z}_{l-1})) \right>_{q_{\boldsymbol{\lambda}}(\mathbf{W}_l, \boldsymbol{\Sigma}_l)}, \\
\nabla_{\mathbf{z}_{l}} \mathcal{E}_{l+1} &= -\mathbf{D}\left(\mathbf{z}_l\right)\left<\mathbf{W}_{l+1}^T \boldsymbol{\Sigma}_{l+1}^{-1} (\mathbf{z}_{l+1} - \mathbf{W}_{l+1} f(\mathbf{z}_{l})) \right>_{q_{\boldsymbol{\lambda}}(\mathbf{W}_{l+1}, \boldsymbol{\Sigma}_{l+1})},
\end{align}

\noindent where \(\mathbf{D}(\mathbf{z}) = \mathrm{diag}(f'(\mathbf{z}))\) is the diagonal Jacobian of the pointwise non-linear transfer function. Under the assumption that \( q_{\boldsymbol{\lambda}}(\mathbf{W}_l, \boldsymbol{\Sigma}_l) \) is Matrix Normal Wishart, the relevant expectations are given by:

\begin{align}
    \left<\boldsymbol{\Sigma}^{-1}_l \mathbf{W}_l\right>_{q_{\boldsymbol{\lambda}}(\mathbf{W}_{l}, \boldsymbol{\Sigma}_{l})} &= \nu_l \boldsymbol{\Psi}_l \mathbf{M}_l, \\
    \left<\mathbf{W}_l^T\boldsymbol{\Sigma}^{-1}_l \mathbf{W}_l\right>_{q_{\boldsymbol{\lambda}}(\mathbf{W}_{l}, \boldsymbol{\Sigma}_{l})} &= \mathbf{M}_l \nu_l \boldsymbol{\Psi}_l \mathbf{M}_l + d_l \mathbf{V}_l.
\end{align}

\noindent or equivalently:

\begin{align}
\nabla_{\mathbf{z}_{l}} \mathcal{E}_{l+1} &= -\mathbf{D}\left(\mathbf{z}_{l}\right)\mathbf{M}^T_{l+1} \left<\boldsymbol{\Sigma}^{-1}_{l+1}\right> (\mathbf{z}_{l+1} - \mathbf{M}_{l+1} f(\mathbf{z}_{l})) \\ 
& \quad + d_{l+1} \mathbf{D}(\mathbf{z}_{l}) \mathbf{V}_{l+1} f(\mathbf{z}_{l}),
\end{align}

\noindent where \( d_l \) is the dimension of \(\mathbf{z}_l\). The storage and compute costs associated with this approximate posterior can be prohibitively expensive, so it may be wise to use structured approximations to \( q_{\boldsymbol{\lambda}}(\mathbf{W}_{l}, \boldsymbol{\Sigma}_{l}) \) consistent with the quadratic form of \(\mathcal{E}(\mathbf{Z}, \boldsymbol{\lambda})\).

\section{Predictive Coding}
\label{sec:pc}

To help clarify the differences between predictive coding (PC) and Bayesian predictive coding (BPC), this section describes PC using the same notation used in Section \ref{sec:methods}.

PC is a method for inverting hierarchical Gaussian generative models with \( L \) layers of variables $\mathbf{Z} = \left\{\mathbf{z}_l: l= 0\dots L\right\}$ and parameters $\boldsymbol{\Theta} = \left\{\mathbf{W}_l,\boldsymbol{\Sigma}_l: l=1\dots L\right\}$:

\begin{equation}
    \begin{aligned}
        p(\mathbf{Z} \mid \boldsymbol{\Theta}) &= p(\mathbf{z}_0) \prod_{l=1}^{L} p(\mathbf{z}_l \mid \mathbf{z}_{l-1}, \mathbf{W}_l,\boldsymbol{\Sigma}_l), \\
        p(\mathbf{z}_l \mid \mathbf{z}_{l-1}, \mathbf{W}_l,\boldsymbol{\Sigma}_l) &= \mathcal{N}(\mathbf{z}_l \mid  \mathbf{W}_l f(\mathbf{z}_{l-1}), \boldsymbol{\Sigma}_l)
    \end{aligned}
    \label{eq:pc_gm}
\end{equation}

In comparison with Equation \ref{eq:joint_distribution}, Equation \ref{eq:pc_gm} does not include priors over parameters $\boldsymbol{\Theta}$. 

In the PC framework, latent variables $\mathbf{Z}$ and parameters $\boldsymbol{\Theta}$ are not given a Bayesian treatment but are instead represented via maximum a posteriori (MAP) and maximum likelihood (ML) estimates, respectively. The estimates are generated via gradient descent, akin to inference in an energy-based model (EBM) where the energy is given by the negative log probability of the model:

\begin{equation}
    \begin{aligned}
        \mathcal{E}(\mathbf{Z}, \boldsymbol{\Theta}) &= - \log \left( p(\mathbf{z}_0) \prod_{l=1}^{L} p(\mathbf{z}_l \mid \mathbf{z}_{l-1}, \mathbf{W}_l, \boldsymbol{\Sigma}_l) \right) \\
        &= \frac{1}{2} \sum_{l=1}^{L} 
        \underbrace{\left(\mathbf{z}_l - \mathbf{W}_l f(\mathbf{z}_{l-1}) \right) \cdot \boldsymbol{\Sigma}^{-1}_l \left(\mathbf{z}_l - \mathbf{W}_l f(\mathbf{z}_{l-1}) \right)}_{\mathcal{E}_l} + C
    \end{aligned}
\label{eq:pc_energy}
\end{equation}

\noindent where $C$ is independent of the $\mathbf{Z}$'s and $\mathcal{E}_l$ is the precision-weighted prediction error. In comparison to Equation \ref{eq:energy_objective}, Equation \ref{eq:pc_energy} is a \textit{function} of the parameters $\boldsymbol{\Theta}$, rather than a function of the posterior parameters $\boldsymbol{\lambda}$. Moreover, the energy function in Equation \ref{eq:pc_energy} does not have an expectation under $q_{\boldsymbol{\lambda}}(\boldsymbol{\Theta})$. As a result, the resulting expression as precision-weighted prediction errors is identical to Equation \ref{eq:energy_rewrite}, modulo the expectation under $q_{\boldsymbol{\lambda}}(\boldsymbol{\Theta})$.

Akin to BPC, PC is also solved using the expectation-maximisation (EM) algorithm, but now directly on the maximum-likelihood estimates of $\boldsymbol{\Theta}$ rather than the variational parameters $\boldsymbol{\lambda}$:

\begin{equation}
    \mathbf{Z}^* = \arg \min_{\mathbf{Z}} \mathcal{E}(\mathbf{Z}, \boldsymbol{\Theta}), \quad \boldsymbol{\Theta}^* = \arg \min_{\boldsymbol{\Theta}} \mathcal{E}(\mathbf{Z}^*, \boldsymbol{\Theta}).
\end{equation}

To solve these optimisation problems, PC relies on gradient descent for both the latent variables and parameters, where the gradients are given by:

\begin{equation}
    \nabla_{\mathbf{z}_l} = \frac{\partial \mathcal{E}}{\partial \mathbf{z}_l} = \frac{1}{2} \left( \frac{\partial \mathcal{E}_l}{\partial \mathbf{z}_l} + \frac{\partial \mathcal{E}_{l+1}}{\partial \mathbf{z}_l} \right), \quad
    \nabla_{\boldsymbol{\theta}_l} = \frac{\partial \mathcal{E}}{\partial \boldsymbol{\theta}_l} = \frac{1}{2} \frac{\partial \mathcal{E}_l}{\partial \boldsymbol{\theta}_l}.
\end{equation}

The resulting update rule for the $\mathbf{Z}$ takes the form:

\begin{equation}
\mathbf{z}_l \rightarrow \mathbf{z}_l -\alpha \left( \boldsymbol{\Sigma}^{-1}_l \mathbf{z}_l - \mathbf{D}(\mathbf{z}_l)  \mathbf{W}_{l+1}^T\boldsymbol{\Sigma}_{l+1}^{-1} (\mathbf{z}_{l+1} - \mathbf{W}_{l+1} \mathbf{f}(\mathbf{z}_l) \right)
\end{equation}

We note that the derivative of the non-linear activation function is bounded by 1 this is a weakly non-linear dynamical system with a block triangular structure. Thus the dynamics dominated by the spectrum of $\mathbf{A}_l = \Sigma^{-1}_l + \mathbf{W}_{l+1}^T\Sigma^{-1}_{l+1}\mathbf{W}_{l+1}$ allowing us to identify an upper bound on the maximum learning rate parameter as approximately given by the inverse of the maximum eigenvalue of the $\mathbf{A}_l$ or the sum of the maximum eigenvalues associated with each component, which can be dynamically updated with updates to the posterior distribution over the parameters.

\section{Conjugate updates}
\label{sec:closed_form}

Here we show that the parameters of $q(\mathbf{W}_l, \boldsymbol{\Sigma}_l)$ that minimize the energy defined in \Cref{eq:energy_objective} can be written as a sum of the sufficient statistics of the likelihood $p(\mathbf{z}_l\mid \mathbf{z}_{l-1}, \mathbf{W}_l, \boldsymbol{\Sigma})$ and the natural parameters of the prior $p(\mathbf{W}_l, \boldsymbol{\Sigma}_l)$. 

Consider the energy function for the $l$\textsuperscript{th} layer of the network, which depends on the $l$ and $l-1$\textsuperscript{th} latent variables and the Matrix Normal Wishart parameters of that layer:

\begin{equation}
        \mathcal{E}(\mathbf{z}_l, \boldsymbol{\lambda}_l, \mathbf{z}_{l-1}) = \Big< \log q_{\boldsymbol{\lambda}}(\mathbf{W}_l, \boldsymbol{\Sigma}_l) - \log p(\mathbf{z}_l, \mathbf{W}_l, \boldsymbol{\Sigma}_l|\mathbf{z}_{l-1}) \Big>_{q_{\boldsymbol{\lambda}}(\mathbf{W}_l, \boldsymbol{\Sigma}_l)} 
        \label{eq:energy_objective_layerwise}
\end{equation}

We obtain the optimal value of the parameters, which we denote $\boldsymbol{\lambda}^{*}_l$, by setting the derivative of this local energy function to 0 and solving for $\boldsymbol{\lambda}_l$. The conjugacy of the likelihood and prior $p(\mathbf{z}_l, \mathbf{W}_l, \boldsymbol{\Sigma}_l|\mathbf{z}_{l-1})$, combined with the Matrix Normal Wishart form of the variational posteriors over $\mathbf{W}_l, \boldsymbol{\Sigma}_l$, means that the variational parameters $\boldsymbol{\lambda}_{l}$ can be equated with the natural parameters of a Matrix Normal Wishart distribution, i.e.,

\begin{align}
    \boldsymbol{\lambda}_l &\equiv \boldsymbol{\eta}_l \notag \\
    \boldsymbol{\eta}_l&= \begin{bmatrix} \mathbf{V}^{-1}_l \\ \mathbf{M}_l \mathbf{V}^{-1}_l \\ \boldsymbol{\Phi}_l + \mathbf{M}_l \mathbf{V}^{-1}_l \mathbf{M}^{\top}_l\\ \nu_l - d_y + d_x - 1 \end{bmatrix}
\end{align}

The value of $\boldsymbol{\eta}_l$ at which the derivative of \Cref{eq:energy_objective_layerwise} vanishes, results in a succinct expression in terms of the natural parameters of the prior $\boldsymbol{\eta}_0$ and the sufficient statistics of the $\mathbf{z}_{l-1}$-conditioned likelihood over $\mathbf{z}_l$:

\begin{align}
    \boldsymbol{\eta}^{*}_l&= \begin{bmatrix} \mathbf{V}^{-1}_{l,0} + f(\mathbf{z}_{l-1}) f(\mathbf{z}_{l-1})^\top \\ \mathbf{M}_{l,0} \mathbf{V}^{-1}_{l,0} + f(\mathbf{z}_{l-1})\mathbf{z}_{l}^\top\\ \boldsymbol{\Phi}_{l,0} + \mathbf{M}_{l,0} \mathbf{V}^{-1}_{l,0} \mathbf{M}^{\top}_{l,0} + \mathbf{z}_{l}\mathbf{z}_{l}^\top\\ \nu_{l,0} - d_y + d_x - 1 + 1\end{bmatrix}
\end{align}

The $\mathbf{X}_{l,0}$ notation denotes parameters of the Matrix Normal Wishart prior for the $l$\textsuperscript{th} layer. Because this model is fully conjugate, this update is an exact Bayesian update and thus the optimal variational posterior is equal to the true posterior \citep{beal2003variational}.

\section{Related work}
\label{sec:related_work}

Several extensions to predictive coding (PC) have previously been proposed. Variational Laplace \citep{zeidman2023primer, friston2007variational} employs a Laplace approximation, resulting in an algorithm expressed in terms of precision-weighted prediction errors. This approach additionally includes priors on model parameters, resulting in maximum \textit{a posteriori} (MAP) estimates. However, it does not represent full posterior distributions over parameters, thus limiting its ability to quantify epistemic uncertainty. Other approaches have explored incorporating sparse priors into the generative model \citep{ororbia2022neural}, but these also do not infer full posterior distributions. Variants of PC employing Monte Carlo sampling and Langevin dynamics \citep{oliviers2024learning, zahid2023sample} have successfully inferred posterior distributions over latent variables, but have not been extended to infer distributions over parameters, again limiting their capacity to capture epistemic uncertainty. An interesting direction for future research would be to integrate our Bayesian predictive coding (BPC) algorithm with such sampling-based approaches, potentially leading to a fully Bayesian PC method that infers posterior distributions over both latent variables and parameters.

\section{Experiment details}
\label{sec:experiment_details}

\subsection{Accuracy}

In this section, we provide additional details regarding the experiments presented in Section \ref{sec:acc}. For all datasets, reported performance is averaged over 5 random seeds. In the regression task (\texttt{energy}), we report performance using the mean squared error (MSE) between the predicted and true values. For classification tasks (\texttt{MNIST} and \texttt{two moons}), we use one-hot encoded labels during training. At test time, the predicted class label is determined by taking the \texttt{argmax} over the output nodes, where the number of nodes corresponds to the number of classes.

For the \texttt{energy} and \texttt{MNIST} datasets, we employ the same neural network architecture. Specifically, we use a four-layer neural network with 128 hidden units per layer and \texttt{ReLU} activations. Training is performed using mini-batches of size 128. We initialized the linear weights \(\mathbf{W}\) of shape (\texttt{out\_features}, \texttt{in\_features}) from a uniform distribution \(\mathcal{U}(-\sqrt{k}, \sqrt{k})\), where \(k = \frac{1}{\texttt{in\_features}}\). Similarly, the bias \(\mathbf{b}\) of shape (\texttt{out\_features}) is initialized from \(\mathcal{U}(-\sqrt{k}, \sqrt{k})\).

For BP, we used the \texttt{Adam} \citep{kingma2014adam} optimizer with a learning rate of 0.001. For PC, we used the \texttt{AdamW} \citep{loshchilov2017decoupled} optimizer for the parameters with a learning rate of $0.0002$ and a weight decay value of $0.65$, and a stochastic gradient descent optimiser for the hidden states with a learning rate of $0.01$ and momentum of $0.65$. These parameter values followed those proposed in \citep{pinchetti2024benchmarking} for the same dataset. We used $10$ iterations of hidden state updates for each batch, before performing one gradient step on the weights. 

For BPC, we used the \texttt{Adam} \citep{kingma2014adam} optimizer for hidden states, with a learning rate of $0.01$ and $10$ iterations per batch. For the parameter learning rate, we used $\kappa_t = t^{-\epsilon}$, where $t$ is the total number of updates and $-\epsilon$ was set to $0.25$. 

We set the prior over the weights $\mathbf{M}^{(0)}$ to be a matrix of zeros of the appropriate size. We set the prior over $\mathbf{V}^{(0)}$ to be $10 \cdot \mathbf{I}$ where $\mathbf{I}$ is an identity matrix of the appropriate size and set $\boldsymbol{\Psi}^{(0)}$ to be $1000 \cdot \mathbf{I}$. Finally, we set $\nu^{(0)}$ to be $d_y + 2$. All initial estimates of the posterior natural parameters $\boldsymbol{\eta}$ were set to the same as the prior, besides $\mathbf{M}$ which uses the initialisation described for $\mathbf{W}$.

The hyper-parameters for the \texttt{two moons} dataset experiments remain the same, except that we use a smaller network architecture consisting of a single hidden layer with 100 hidden units.

\subsection{Synthetic regression}

In the synthetic regression tasks, we use the same parameter settings used in \texttt{two moons}. For the aleatoric uncertainty plot, we generate samples from:

\[
\mathbf{y} = -(\mathbf{x} + 0.5) \cdot \sin(3\pi \mathbf{x}) + \mathcal{N}\left(0, \, (0.45 \cdot (\mathbf{x} + 0.5))^2\right),
\]

where \(\mathbf{x}\) is normally distributed around zero. For the epistemic uncertainty plot, we use:

\[
\mathbf{y} = \mathbf{x}^3 + \mathcal{N}(0, 9),
\]

where $\mathbf{x}$ is uniformly sampled with half of the values drawn from the interval $[3, 5]$ and the other half from $[-5, -3]$.

\subsection{UCI datasets}

For the UCI datasets, we use the same parameter settings as in previous experiments, but with networks that have two hidden layers and $50$ hidden nodes. For Bayes by Backprop (BBB) \cite{blundell2015weight}, we use the \texttt{Adam} \citep{kingma2014adam} optimizer with a learning rate of $0.001$, a prior mean of $0.0$, a $\sigma$ of $1.0$, and a batch size of 100. When computing the log predictive density (LPD) for both BBB and BPC, we draw 20 posterior samples per data batch.